\documentclass{article}

\usepackage[utf8]{inputenc} 
\usepackage[T1]{fontenc}    
\usepackage{hyperref}       
\usepackage{url}            
\usepackage{booktabs}       
\usepackage{amsfonts}       
\usepackage{nicefrac}       
\usepackage{microtype}      
\usepackage{xcolor}         

\usepackage{amsmath,amsfonts,amssymb,latexsym,epsfig}
\usepackage{mathrsfs,mdframed, float,graphicx,dsfont,wrapfig}
\usepackage{mathrsfs}
\usepackage{verbatim}
\usepackage{latexsym}
\usepackage{amsthm}
\usepackage{amssymb}
\usepackage{graphics}
\usepackage{amsbsy}
\usepackage{enumerate}
\usepackage{times}
\usepackage{mathtools}   
\mathtoolsset{showonlyrefs=true}  
\usepackage{bm}
\usepackage{sidecap}
\usepackage{cleveref}
\usepackage{wrapfig}

\definecolor{brinkpink}{rgb}{0.98, 0.38, 0.5}
\definecolor{blue_dark}{rgb}{0.0, 0.5, 0.69}

\title{On original and latent space connectivity in deep neural networks}

\author{%
  Boyang Gu \\
  Imperial College London\\
  \texttt{boyang.gu19@imperial.ac.uk} \\
  Anastasia Borovykh$^*$ \\
  Imperial College London\\
  \texttt{a.borovykh@imperial.ac.uk} \\
}

\begin{document}

\maketitle

\begin{abstract}
We study whether inputs from the same class can be connected by a continuous path, in original or latent representation space, such that all points on the path are mapped \emph{by the neural network model} to the same class. Understanding how the neural network views its own input space and how the latent spaces are structured has value for explainability and robustness. We show that paths, linear or nonlinear, connecting same-class inputs exist in all cases studied. 
\end{abstract}

\section{Introduction}
\begin{wrapfigure}{R}{0.4\textwidth}
    \includegraphics[width=0.4\textwidth]{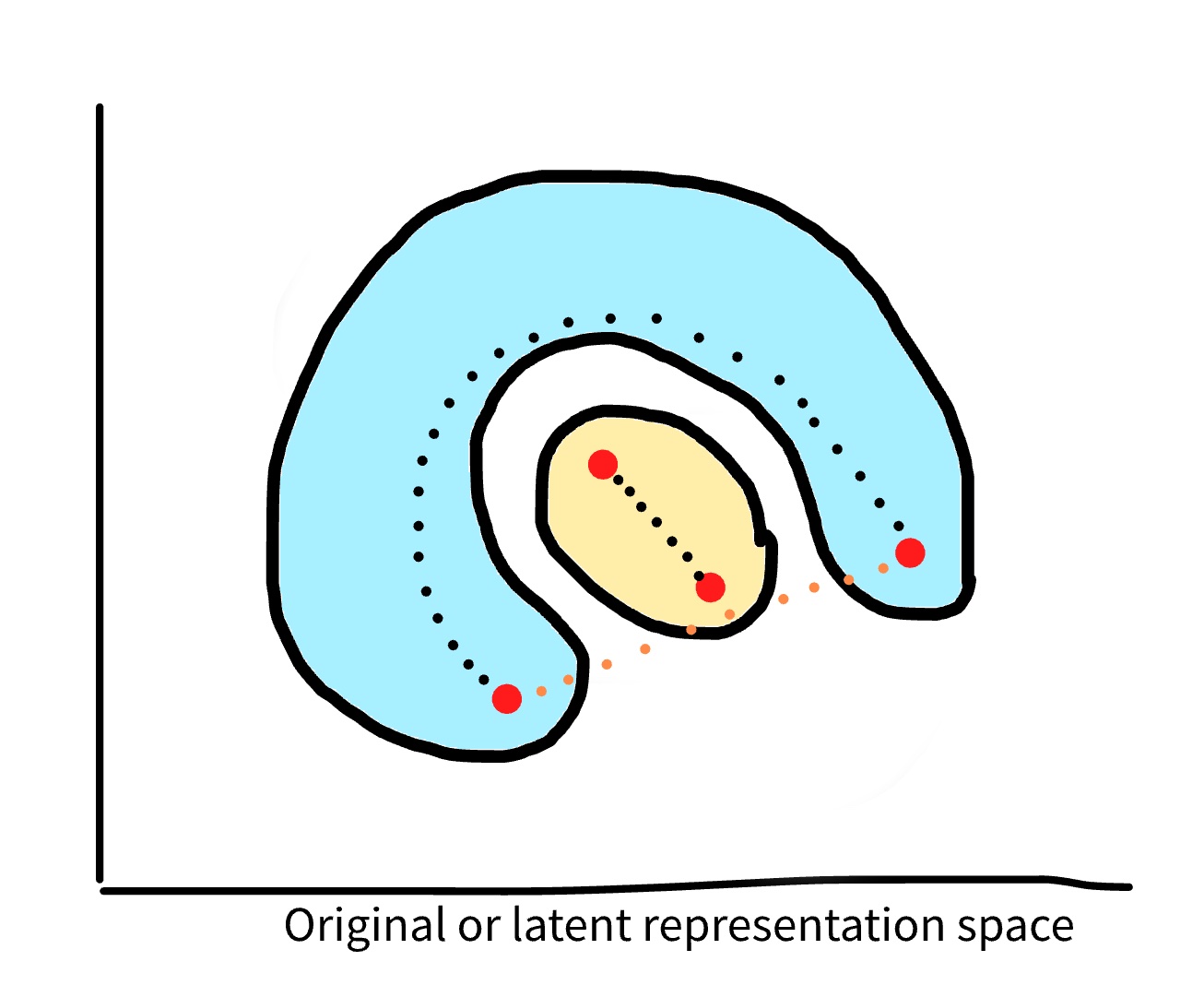}\vspace{-.5cm}
    \caption{In this work we aim to answer an easier question: given two datapoints belonging to a certain target class, can we find paths between these datapoints such that all datapoints on that path belong to the target class too?}
    \label{fig:fig1}
\end{wrapfigure}

The manifold hypothesis states that high-dimensional real-world data typically lies in a lower-dimensional submanifold, the axes of this dimensionality-reduced space representing factors of variation \cite{tenenbaum1997mapping,fefferman2016testing}. Relatedly, the flattening hypothesis \cite{brahma2015deep} and work in disentanglement \cite{poole2016exponential} states that througout learning, subsequent layers in a deep neural network (DNN) disentangle the data in such a way that finally a linear model can separate the classes. Understanding how a DNN itself views its input space can be related to explainability (e.g. \cite{ahern2019normlime}), the models' generalisation (e.g. \cite{kawaguchi2017generalization, chung2018classification}) and sensitivity to adversarial attacks (e.g. \cite{szegedy2013intriguing}). In this work we aim to better understand the structure of the original and latent space representations for a trained neural network and in particular we ask the following question: \emph{Are inputs from the same class connected in original or latent representation space by a path on which every point belongs to the same class?} Figure \ref{fig:fig1} illustrates this question. We stress that we aim to answer this question from a neural networks' viewpoint (i.e. does the model consider these points to belong to the same class). 

Data reconstruction methods have shown that from the model parameters it is possible to reconstruct inputs that are exact copies of datapoints from the train dataset (e.g. \cite{haim2022reconstructing}) or inputs that capture the most common features of a certain class $y$ (e.g. \cite{mahendran2015understanding}). However, in order to reconstruct something sensible, strong additional priors are added into the loss function; purely considering an objective of the form $\min_{x\in\mathcal{X}}l(y-f(x|\theta))$ (where $\mathcal{X}$ is the original input space, $l$ is some error metric $l$, $f$ is the neural network parameterised by $\theta$) will typically result in noise (e.g. consider \cite{olah2017feature} and references therein). Note furthermore that neural networks are known to be sensitive to adversarial perturbations - perturbations consisting of a very small magnitude of random noise that drastically change the classification of the input. This naturally leads to the question of how exactly neural networks view their input space, i.e. how is the input region divided into different classes? 

Instead of working directly in input space one can ask similar questions for the representations in latent space. One could expect that representations for inputs from the same class would lie close together, this closeness increasing as we progress through the depth of the network\cite{brahma2015deep,recanatesi2019dimensionality,magai2023deep}. However, it is not clear whether \emph{all} representations from the same class will lie in a connected subspace of the representation space. Gaining understanding into this is the main objective of this work. 

\section{Related work}

\textbf{The geometry of the input and latent space}
The manifold hypothesis states that high-dimensional real-world data typically lies in a much lower-dimensional submanifold which preserves the intrinsic metric structure of the observations \cite{tenenbaum1997mapping}. Considering the viewpoint of neural networks, work such as \cite{brahma2015deep,recanatesi2019dimensionality,magai2023deep} explores how the manifold unfolds throughout learning. Other work shows that DNNs divide the input
space into linear regions in each of which the DNN behaves as a linear function \cite{montufar2014number, serra2018bounding, tiwari2022effects}. We extend upon this piecewise linear separation by asking whether points from the same class can be connected with a linear or nonlinear path of same-class points in both original and latent representation space (and not just in the last layer).
\vspace{0.2cm}\\
\textbf{Latent space understanding}
The structure of the latent space can also be used to understand e.g. model failures \cite{jain2022distilling}, generate outputs with particular features \cite{parihar2022everything} or compare representations between layers or models \cite{kornblith2019similarity,morcos2018insights}. Alternatively, the work of \cite{goldt2020modeling} uses the (lower-dimensional) latent space to generate high-dimensional inputs with labels that depend only on their position within this latent space. They note that most points in the high-dimensional space cannot be interpreted as images (they are seemingly pure noise) and only those points on the lower-dimensional manifold can be seen as real images; this would relate to the observation that simply optimising in input space for a particular class outcome results most often in pure noise. 
\vspace{0.2cm}\\
\textbf{Connectivity in weight space} Prior work \cite{draxler2018essentially, garipov2018loss, freeman2016topology} has shown that the loss landscape of a neural network (as a function of its weights) consists of connected valleys, all points in this valley attaining equally low loss values. More recently \cite{frankle2020linear} showed that models optimised under differing SGD noise, but starting from the same initialisation, can be connected through a \emph{linear} path. We ask a similar question but work in the input and latent representation space.

\section{Preliminaries and methodology}
We consider deep neural networks trained for classification using a cross-entropy loss. Our goal is to understand the structure of the input space through the lens of the neural network itself. More specifically, we ask whether we can find a path between two datapoints of the same class $y$ such that all points on this path are classified (by the DNN) to the belong to $y$. 

We begin with two inputs $\mathbf{x}_1,\mathbf{x}_2$ whose true labels are from the same class. When finding paths in original representation space we work directly in this space aiming to find a path between $\mathbf{x}_1$ and $\mathbf{x}_2$. When finding paths in latent representation space we map the inputs $\mathbf{x}_1,\mathbf{x}_2$ into their `latent representations' given by the outputs $\mathbf{z}_1^l,\mathbf{z}_2^l$ from hidden layer $l$ and aim to find a path in this space. 
\vspace{0.2cm}\\
\textbf{Linear interpolation}
Our baseline for understanding the connectivity structure consists of a linear interpolation between either $\mathbf{x}_1,\mathbf{x}_2$ or $\mathbf{z}_1^l,\mathbf{z}_2^l$. If all points on this linear interpolation belong to the same class (according to the neural network classifier) we say that the datapoints are \emph{linearly connectable}. 
\vspace{0.2cm}\\
\textbf{Nudged elastic band method}
For ease of notation we use $\mathbf{z}_1^l,\mathbf{z}_2^l$ with $l=0$ to represent the original inputs $\mathbf{x}_1,\mathbf{x}_2$. The goal is to find a continuous path $p^*$ from $\mathbf{z}_1^l$ to $\mathbf{z}_2^l$ through this representation space (original or latent) with the lowest loss,
\begin{align}
    p(\mathbf{z}_1^l,\mathbf{z}_2^l)^* = \arg\min_{p\in\mathcal{P}}\max_{\mathbf{z}\in p}L(\mathbf{z}),
\end{align}
where $\mathcal{P}$ denotes all possible paths between $\mathbf{z}_1^l$ to $\mathbf{z}_2^l$. We refer to this path as the minimum energy path (MEP). In order to optimise for all points on the path being classified to the same class we use $L(\mathbf{z}) = l(y,f_{l:L}(\mathbf{z}^l|\theta))$ where $f_{l:L}(\cdot|\theta)$ denotes the \emph{partial} neural network that maps $\mathbf{z}^l$ to the output, $y$ is the one-hot encoded \emph{target} class and $l$ the cross-entropy loss. To find this path we use the Nudged Elastic Band (NEB) method \cite{jonsson1998nudged}. Starting from a straight line with $N$ pivot points (our linear interpolation with $N$ intermediate points), we apply gradient forces iteratively to minimise the loss among the path. We refer to \cite{jonsson1998nudged} or \cite{draxler2018essentially} for a detailed description of the method, but note that the energy function for the path consists of the original loss $L$ and a spring force loss that aims to ensure some amount of equidistance between the points. 

\paragraph{Decoding the latent space}
In order to have a better interpretation of the points on the minimum energy path in latent space, we train a decoder in the style of \cite{gopalakrishnan2022classify} that maps the latent space representations back into the original input space. The structures of the decoders are shown in Table~\ref{tab:veryfirst_decoder} and Table~\ref{tab:last_decoder} in Appendix \ref{app:app1}. For training the decoders, we use Adam with hyperparameters as explained in Table~\ref{tab:decoder_hyper} in Appendix \ref{app:app1}.

\section{Results}
We consider a ResNet50 model pretrained on ImageNet-10K and finetuned on CIFAR10 using the cross-entropy loss and Adam. Our goal is to find MEPs in original input and latent representation space. Our main results are presented in Table \ref{table:main_results}. 

\begin{table}
  \caption{For a ResNet50 DNN finetuned on CIFAR10 we compute the number of times (out of 25 different pairs of points $\mathbf{z}^l_1,\mathbf{z}_2^l$) i) a linear path exists on which all points belong to the same class (as classified by the DNN), ii) a nonlinear path exists on which all points belong to the same class, iii) no such path exists.}
  \label{table:main_results}
  \centering
  \begin{tabular}{llll}
    \toprule
    & Original & Last layer & Intermediate layer\\
    \hline\hline
    Linear path exists &  19 & 25 & 14\\
    Nonlinear path exists & 6 & 0 & 11\\
    No path exists &  0 & 0 & 0 \\
    \bottomrule
  \end{tabular}
\end{table}

The majority of pairs of datapoints can be connected by a linear path in both the original input space as well as the latent representation spaces. While this is to be expected for the last layer (through the flattening hypothesis and prior works on piecewise linear separation\cite{montufar2014number, serra2018bounding, tiwari2022effects}), it is intriguing that also the original input space seems to be consisting of regions inside which all points belong to the same class. 

In Figure \ref{fig:fig_nonlinear} we present examples in which the points on a linearly interpolated path are classified to different classes, however optimising for the minimum energy path leads (after a very small number of iterations) to a nonlinear path on which all points do belong to the target class. More specifically, for the original input space there was no linear path connecting the two images. However, for the last layer, a linear path connecting the two does exist. This strengthens the idea that depth in a neural network disentangles the input space in such a way that representations are easier to separate. 

Finally, we note that the first intermediate layer contains less linearly connectable datapoints compared to both original input space and the last layer representation. Noteworthy is that visually, the intermediate images that are `misclassified' do not differ strongly from the correctly classified images (those after optimisation for the MEP); this can be related to the idea that small perturbations of the input can drastically change the classification of it. 

\begin{figure}[h!]
    \centering
    \includegraphics[trim={3cm 0 1cm 4cm},clip, width=0.25\textwidth]{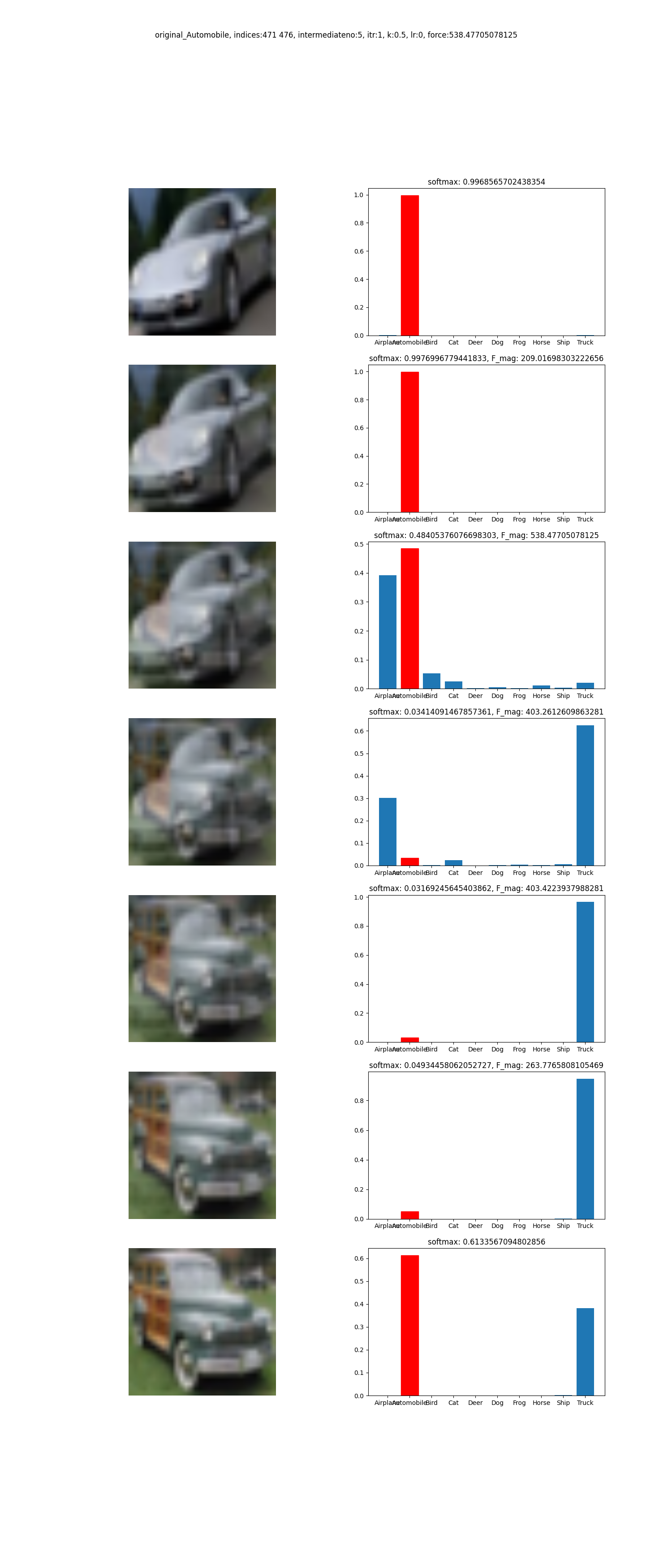}\hspace{-0.5cm}
    \includegraphics[trim={3cm 0 1cm 4cm},clip, width=0.25\textwidth]{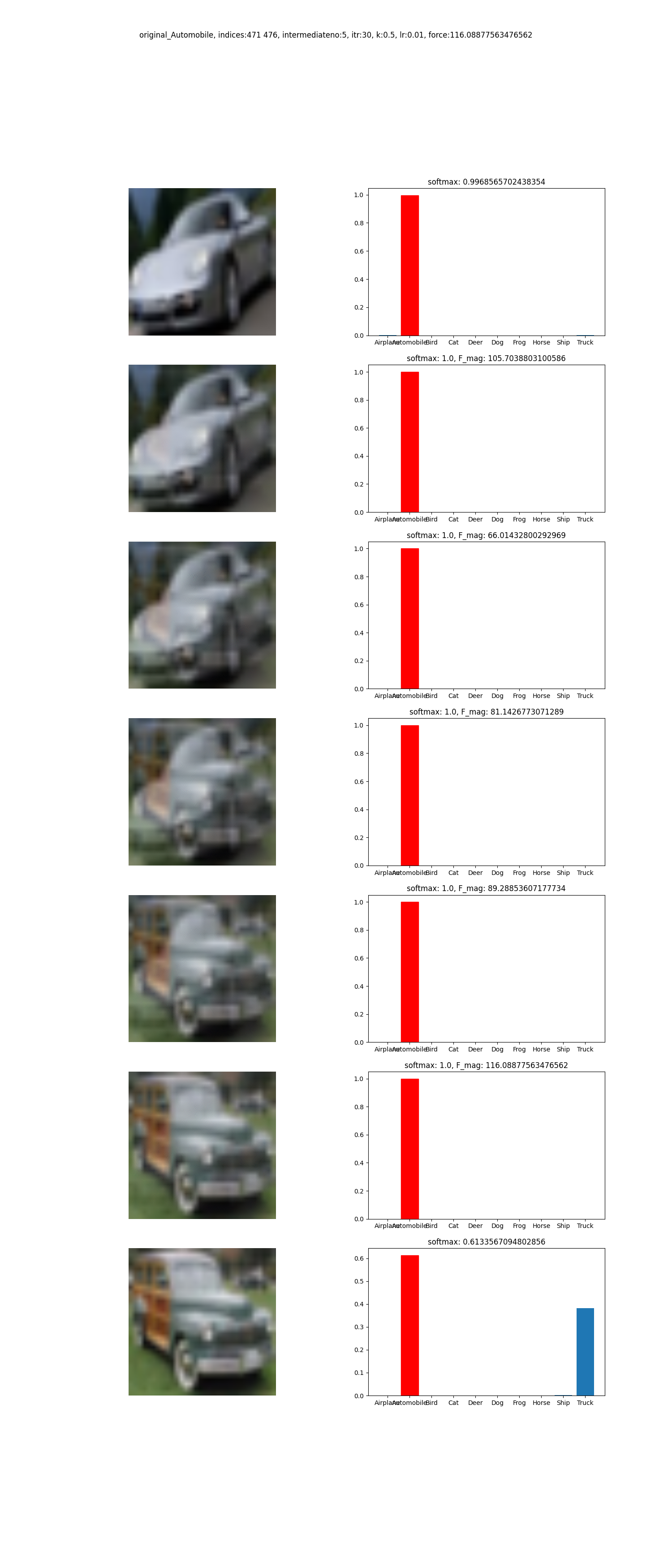}\hspace{-0.5cm}
    \includegraphics[trim={3cm 0 1cm 4cm},clip, width=0.25\textwidth]{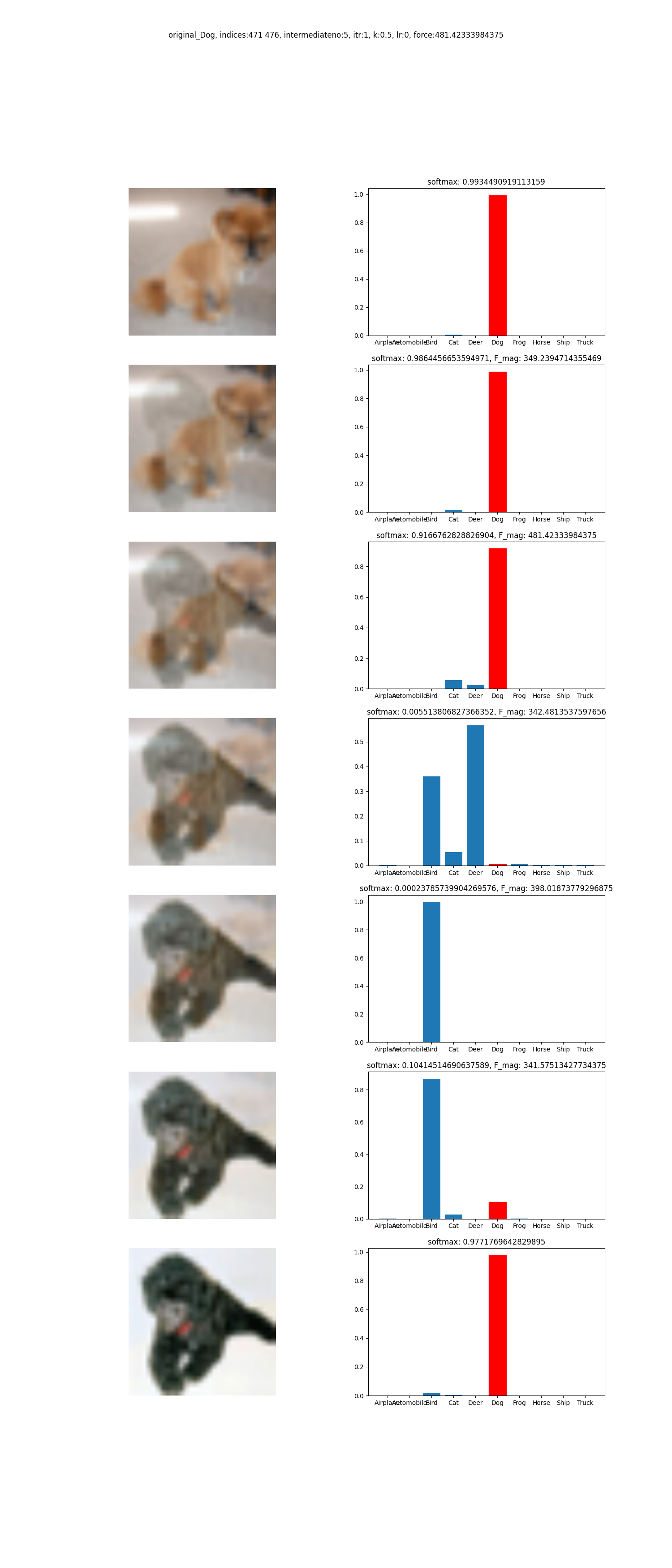}\hspace{-0.5cm}
    \includegraphics[trim={3cm 0 1cm 4cm},clip, width=0.25\textwidth]{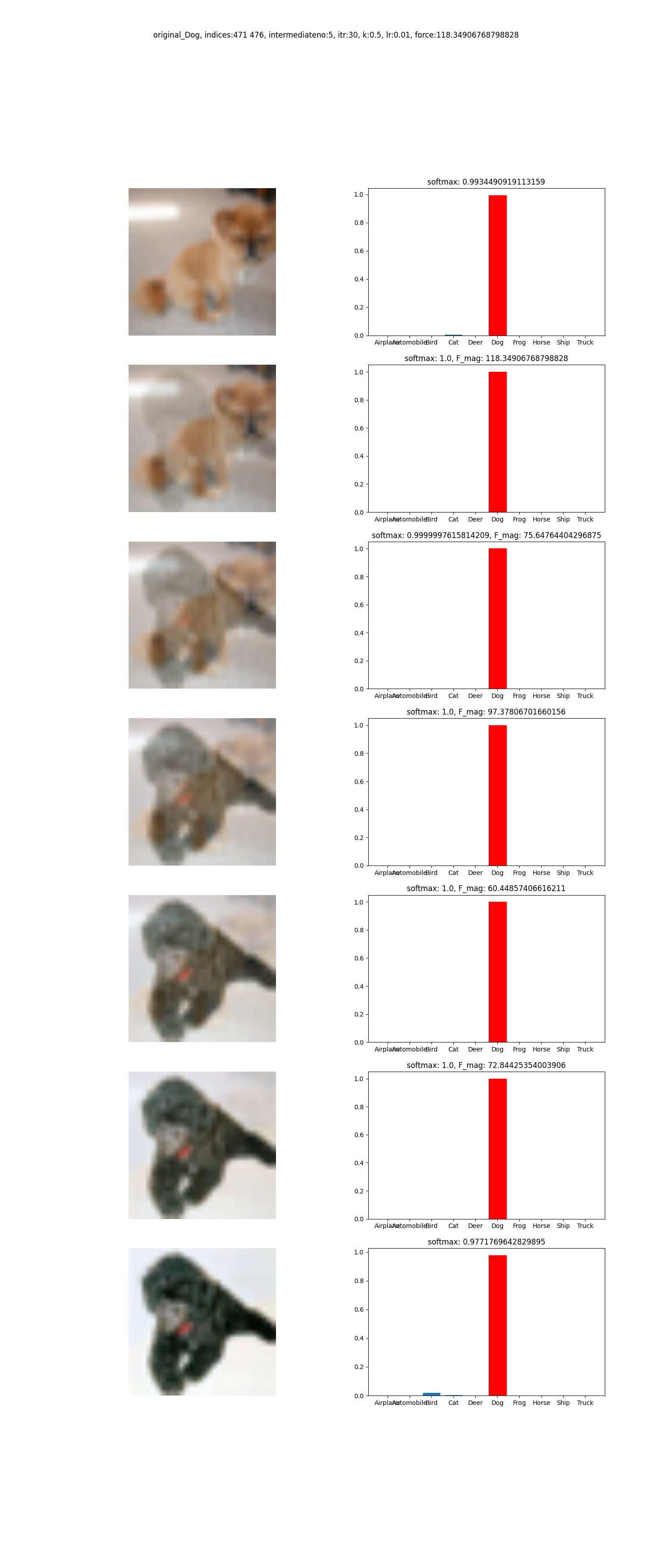}
    \vspace{-1cm}
    \caption{Results for the \emph{original} input space; the bars are the softmax probabilities per class with red denoting the target class. Two left-most images and two right-most images are both settings in which only a nonlinear path connecting $\mathbf{z}^0_1$ and $\mathbf{z}^0_2$ exists; each (L) one is a linear interpolation in which some points are misclassified , while each (R) one is the nonlinear MEP in which the target class is attained for each input. }
    \label{fig:fig_nonlinear}
\end{figure}

\begin{figure}[h!]
    \centering
    \includegraphics[trim={0 0 0 4cm},clip, width=0.25\textwidth]{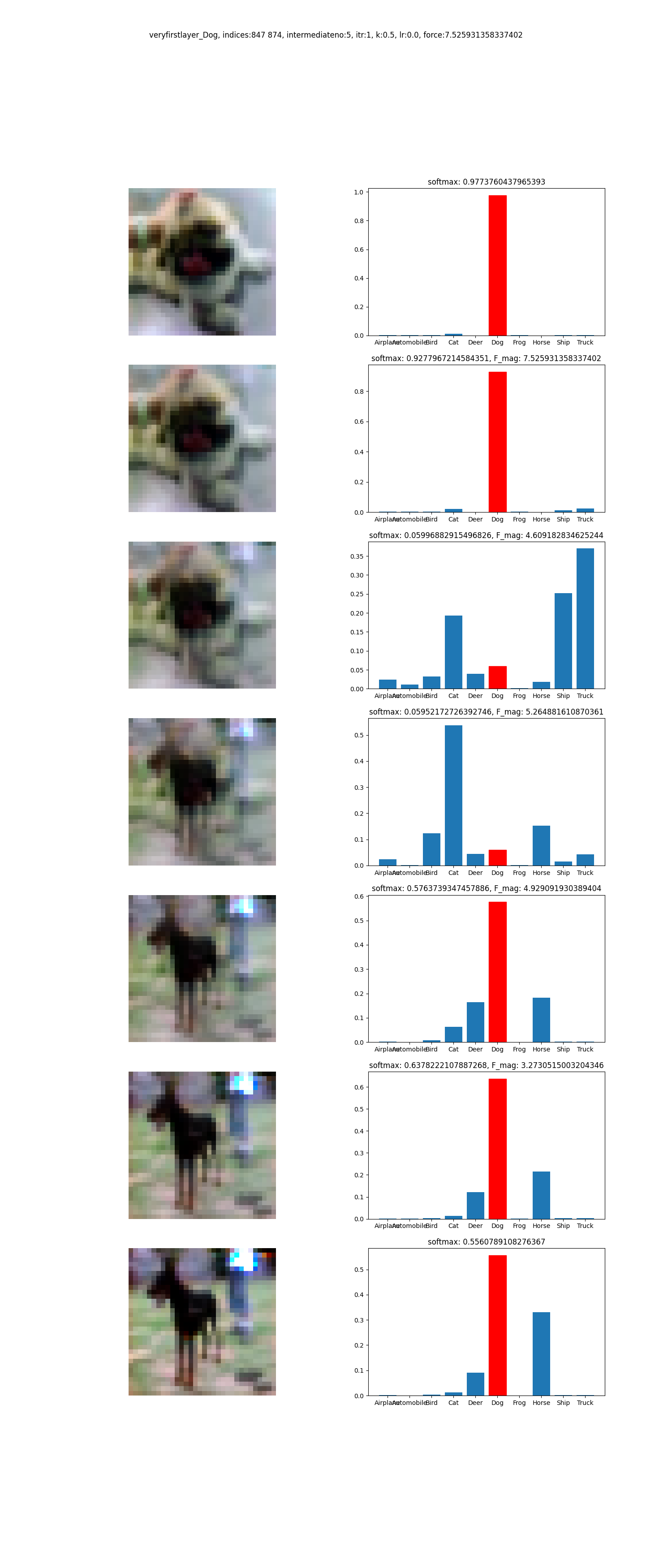}\hspace{-0.5cm}
    \includegraphics[trim={0 0 0 4cm},clip, width=0.25\textwidth]{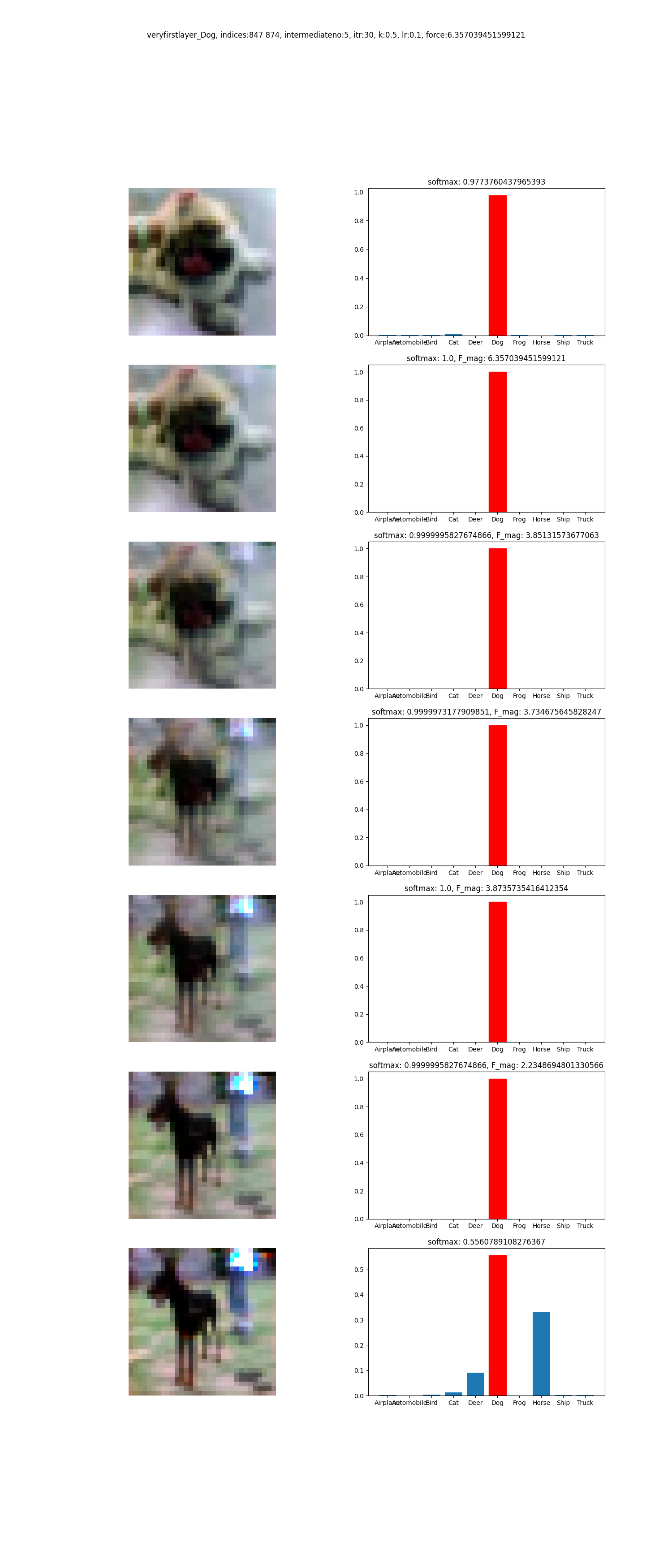}\hspace{-0.5cm}
    \includegraphics[trim={0 0 0 4cm},clip, width=0.25\textwidth]{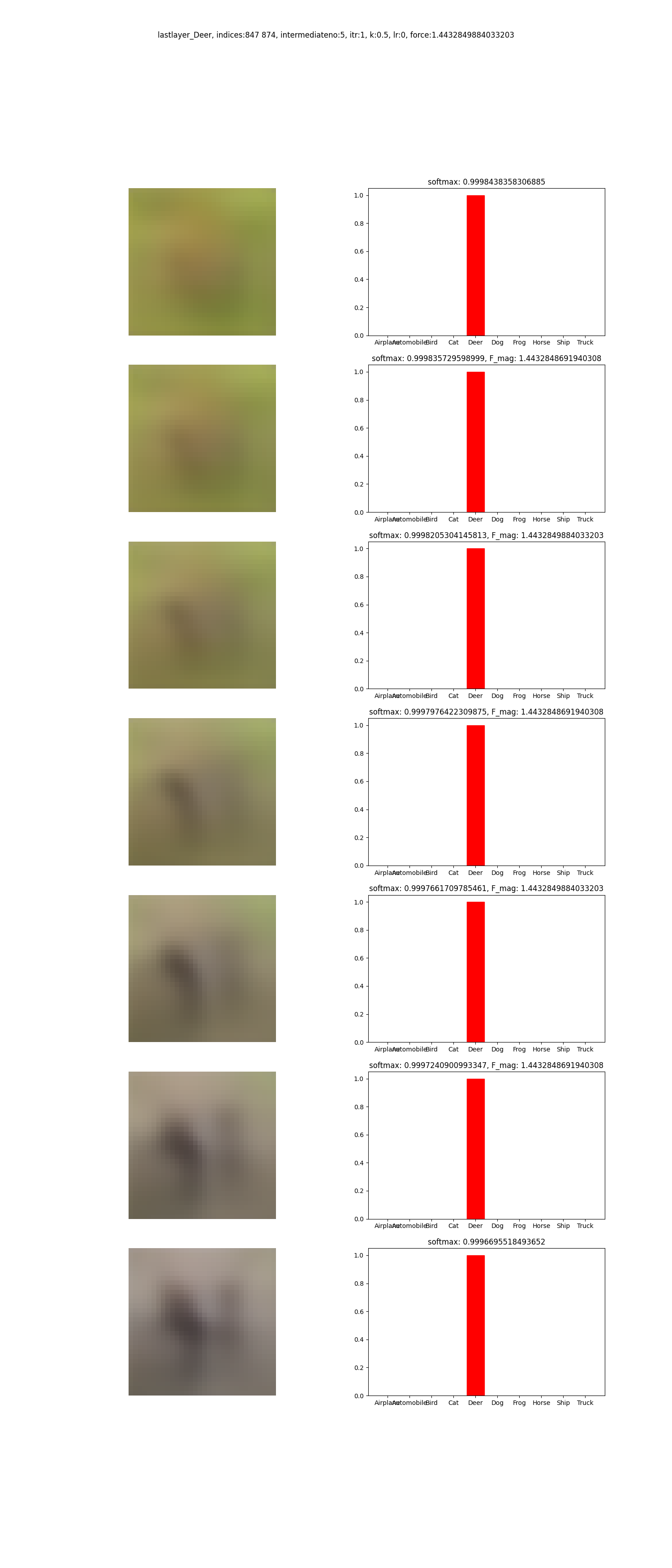}
    \vspace{-1cm}
    \caption{The bars are the softmax probabilities per class with red denoting the correct class (R) and (C) Results for the \emph{latent representation space of the first layer}: a setting in which only a nonlinear path connecting $\mathbf{z}^0_1$ and $\mathbf{z}^0_2$ exists; (R) is a linear interpolation in which some points are classified as cats, while (C) is the nonlinear MEP in which the target class is attained for each input. (L) Results for the \emph{latent representation space of the last layer}: a setting in which a linear path connecting $\mathbf{z}^0_1$ and $\mathbf{z}^0_2$ exists. }
    \label{fig:fig_nonlinear}
\end{figure}

\section{Conclusion and discussion}
In this work we aimed at understanding better what points in original or latent representation space a DNN considers to be part of a particular class. In order to gain insight into this we considered an easier question: can any pair of datapoints be connected by a path all points on which belong to the same class? For all datapoints considered the answer to this question is yes. In order to draw strong conclusions about the original and latent representation space additional experiments are needed: i) one can ask whether for every pair of points belonging to a certain class a connected path exists - hence answering whether disconnected subspaces of same classes exist, ii) one can study the decoded input representations of the points on the connecting path to understand whether the DNN leverages certain common features in order to classify inputs to be of a certain class, iii) one could identify misclassified points and see what their connectivity properties are with other points in the data, iv) one can study the connectivity of representations between different initialisations, runs of the optimiser or different datasets. 

\newpage
\bibliographystyle{plain}
\bibliography{references.bib}

\begin{thebibliography}{10}

\bibitem{ahern2019normlime}
Isaac Ahern, Adam Noack, Luis Guzman-Nateras, Dejing Dou, Boyang Li, and Jun Huan.
\newblock Normlime: A new feature importance metric for explaining deep neural networks.
\newblock {\em arXiv preprint arXiv:1909.04200}, 2019.

\bibitem{brahma2015deep}
Pratik~Prabhanjan Brahma, Dapeng Wu, and Yiyuan She.
\newblock Why deep learning works: A manifold disentanglement perspective.
\newblock {\em IEEE transactions on neural networks and learning systems}, 27(10):1997--2008, 2015.

\bibitem{chung2018classification}
SueYeon Chung, Daniel~D Lee, and Haim Sompolinsky.
\newblock Classification and geometry of general perceptual manifolds.
\newblock {\em Physical Review X}, 8(3):031003, 2018.

\bibitem{draxler2018essentially}
Felix Draxler, Kambis Veschgini, Manfred Salmhofer, and Fred Hamprecht.
\newblock Essentially no barriers in neural network energy landscape.
\newblock In {\em International conference on machine learning}, pages 1309--1318. PMLR, 2018.

\bibitem{fefferman2016testing}
Charles Fefferman, Sanjoy Mitter, and Hariharan Narayanan.
\newblock Testing the manifold hypothesis.
\newblock {\em Journal of the American Mathematical Society}, 29(4):983--1049, 2016.

\bibitem{frankle2020linear}
Jonathan Frankle, Gintare~Karolina Dziugaite, Daniel Roy, and Michael Carbin.
\newblock Linear mode connectivity and the lottery ticket hypothesis.
\newblock In {\em International Conference on Machine Learning}, pages 3259--3269. PMLR, 2020.

\bibitem{freeman2016topology}
C~Daniel Freeman and Joan Bruna.
\newblock Topology and geometry of half-rectified network optimization.
\newblock {\em arXiv preprint arXiv:1611.01540}, 2016.

\bibitem{garipov2018loss}
Timur Garipov, Pavel Izmailov, Dmitrii Podoprikhin, Dmitry~P Vetrov, and Andrew~G Wilson.
\newblock Loss surfaces, mode connectivity, and fast ensembling of dnns.
\newblock {\em Advances in neural information processing systems}, 31, 2018.

\bibitem{goldt2020modeling}
Sebastian Goldt, Marc M{\'e}zard, Florent Krzakala, and Lenka Zdeborov{\'a}.
\newblock Modeling the influence of data structure on learning in neural networks: The hidden manifold model.
\newblock {\em Physical Review X}, 10(4):041044, 2020.

\bibitem{gopalakrishnan2022classify}
Saisubramaniam Gopalakrishnan, Pranshu~Ranjan Singh, Yasin Yazici, Chuan-Sheng Foo, Vijay Chandrasekhar, and ArulMurugan Ambikapathi.
\newblock Classify and generate: Using classification latent space representations for image generations.
\newblock {\em Neurocomputing}, 471:296--334, 2022.

\bibitem{haim2022reconstructing}
Niv Haim, Gal Vardi, Gilad Yehudai, Ohad Shamir, and Michal Irani.
\newblock Reconstructing training data from trained neural networks.
\newblock {\em Advances in Neural Information Processing Systems}, 35:22911--22924, 2022.

\bibitem{jain2022distilling}
Saachi Jain, Hannah Lawrence, Ankur Moitra, and Aleksander Madry.
\newblock Distilling model failures as directions in latent space.
\newblock {\em arXiv preprint arXiv:2206.14754}, 2022.

\bibitem{jonsson1998nudged}
Hannes J{\'o}nsson, Greg Mills, and Karsten~W Jacobsen.
\newblock Nudged elastic band method for finding minimum energy paths of transitions.
\newblock In {\em Classical and quantum dynamics in condensed phase simulations}, pages 385--404. World Scientific, 1998.

\bibitem{kawaguchi2017generalization}
Kenji Kawaguchi, Leslie~Pack Kaelbling, and Yoshua Bengio.
\newblock Generalization in deep learning.
\newblock {\em arXiv preprint arXiv:1710.05468}, 1(8), 2017.

\bibitem{kornblith2019similarity}
Simon Kornblith, Mohammad Norouzi, Honglak Lee, and Geoffrey Hinton.
\newblock Similarity of neural network representations revisited.
\newblock In {\em International conference on machine learning}, pages 3519--3529. PMLR, 2019.

\bibitem{magai2023deep}
German Magai.
\newblock Deep neural networks architectures from the perspective of manifold learning.
\newblock {\em arXiv preprint arXiv:2306.03406}, 2023.

\bibitem{mahendran2015understanding}
Aravindh Mahendran and Andrea Vedaldi.
\newblock Understanding deep image representations by inverting them.
\newblock In {\em Proceedings of the IEEE conference on computer vision and pattern recognition}, pages 5188--5196, 2015.

\bibitem{montufar2014number}
Guido~F Montufar, Razvan Pascanu, Kyunghyun Cho, and Yoshua Bengio.
\newblock On the number of linear regions of deep neural networks.
\newblock {\em Advances in neural information processing systems}, 27, 2014.

\bibitem{morcos2018insights}
Ari Morcos, Maithra Raghu, and Samy Bengio.
\newblock Insights on representational similarity in neural networks with canonical correlation.
\newblock {\em Advances in neural information processing systems}, 31, 2018.

\bibitem{olah2017feature}
Chris Olah, Alexander Mordvintsev, and Ludwig Schubert.
\newblock Feature visualization.
\newblock {\em Distill}, 2017.
\newblock https://distill.pub/2017/feature-visualization.

\bibitem{parihar2022everything}
Rishubh Parihar, Ankit Dhiman, and Tejan Karmali.
\newblock Everything is there in latent space: Attribute editing and attribute style manipulation by stylegan latent space exploration.
\newblock In {\em Proceedings of the 30th ACM International Conference on Multimedia}, pages 1828--1836, 2022.

\bibitem{poole2016exponential}
Ben Poole, Subhaneil Lahiri, Maithra Raghu, Jascha Sohl-Dickstein, and Surya Ganguli.
\newblock Exponential expressivity in deep neural networks through transient chaos.
\newblock {\em Advances in neural information processing systems}, 29, 2016.

\bibitem{recanatesi2019dimensionality}
Stefano Recanatesi, Matthew Farrell, Madhu Advani, Timothy Moore, Guillaume Lajoie, and Eric Shea-Brown.
\newblock Dimensionality compression and expansion in deep neural networks.
\newblock {\em arXiv preprint arXiv:1906.00443}, 2019.

\bibitem{serra2018bounding}
Thiago Serra, Christian Tjandraatmadja, and Srikumar Ramalingam.
\newblock Bounding and counting linear regions of deep neural networks.
\newblock In {\em International Conference on Machine Learning}, pages 4558--4566. PMLR, 2018.

\bibitem{szegedy2013intriguing}
Christian Szegedy, Wojciech Zaremba, Ilya Sutskever, Joan Bruna, Dumitru Erhan, Ian Goodfellow, and Rob Fergus.
\newblock Intriguing properties of neural networks.
\newblock {\em arXiv preprint arXiv:1312.6199}, 2013.

\bibitem{tenenbaum1997mapping}
Joshua Tenenbaum.
\newblock Mapping a manifold of perceptual observations.
\newblock {\em Advances in neural information processing systems}, 10, 1997.

\bibitem{tiwari2022effects}
Saket Tiwari and George Konidaris.
\newblock Effects of data geometry in early deep learning.
\newblock {\em Advances in Neural Information Processing Systems}, 35:30099--30113, 2022.

\end{thebibliography}

\appendix
\section{Hyperparameters}\label{app:app1}
\begin{table}[H]
\centering
\begin{tabular}{c}
\toprule
{Input(shape=(256,56,56)),}\\
{[Conv2DTranspose(filters=256, kernel size=5x5)},\\
{ReLU, BatchNorm,}\\
{Conv2DTranspose(filters=256, kernel size=3x3, padding=1)},\\
{ReLU, BatchNorm,}\\
{Conv2DTranspose(filters=256, kernel size=3x3, padding=1)] x 2},\\
{[Conv2DTranspose(filters=256, kernel size=3x3, padding=1),}\\
{ReLU, BatchNorm] x 3}\\
{[Conv2DTranspose(filters=256, kernel size=3x3, padding=1),}\\
{ReLU, BatchNorm] x 3}\\
{[Conv2DTranspose(filters=256, kernel size=3x3, padding=1),}\\
{ReLU, BatchNorm] x 3}\\
{Conv2D(filters=3, kernel size=3x3, padding=1, stride=2), Sigmoid Activation}\\
\bottomrule
\end{tabular}
\caption{Decoder architecture details for the first hidden layer.}
\label{tab:veryfirst_decoder}
\end{table}
\begin{table}[H]
\centering
\begin{tabular}{c}
\toprule
{Input(shape=(2048)), Reshape(shape=(128,4,4))}\\
{[Conv2DTranspose(filters=512, kernel size=3x3, padding=1)},\\
{LeakyReLU(a=0.3), BatchNorm] x 3, UpSampling2D(size=2x2)}\\
{[Conv2DTranspose(filters=256, kernel size=3x3, padding=1),}\\
{LeakyReLU(a=0.3), BatchNorm] x 3, UpSampling2D(size=2x2)}\\
{[Conv2DTranspose(filters=256, kernel size=3x3, padding=1),}\\
{LeakyReLU(a=0.3), BatchNorm] x 3, UpSampling2D(size=2x2)}\\
{[Conv2DTranspose(filters=256, kernel size=3x3, padding=1),}\\
{LeakyReLU(a=0.3), BatchNorm] x 3}\\
{Conv2D(filters=3, kernel size=3x3, padding=1), Sigmoid Activation}\\
\bottomrule
\end{tabular}
\caption{Decoder architecture detail for the last layer.}
\label{tab:last_decoder}
\end{table}
\begin{table}[H]
\centering
\begin{tabular}{ccc}
\toprule
& First hidden layer & Last layer\\
\hline\hline
batch size & 32 & 128\\
epoch & 10 & 5\\
learning rate & 0.001 & 0.001\\
loss& MSE & MSE\\
\bottomrule
\end{tabular}
\caption{Decoder architecture training details.}
\label{tab:decoder_hyper}
\end{table}
\end{document}